# GSG: A Granary Soft Gripper with Mechanical Force Sensing via 3-Dimensional Snap-Through Structure

Huixu Dong, Chao-Yu Chen, Chen Qiu, Chen-Hua Yeow, Haoyong Yu

*Abstract*— Grasping is an essential capability for most robots in practical applications. Soft robotic grippers are considered as a critical part of robotic grasping and have attracted considerable attention in terms of the advantages of the high compliance and robustness to variance in object geometry; however, they are still limited by the corresponding sensing capabilities and actuation mechanisms. We propose a novel soft gripper that looks like a 'granary' with a compliant snap-through bistable mechanism fabricated by integrated mold technology, achieving 'sensing' and 'actuation' purely mechanically. In particular, the snap-through bistable structure in the proposed gripper allows us to reduce the complexity of the mechanism, control, sensing designs since the grasping and sensing behaviors are completely passive. The grasping behaviors are automatically motivated once the trigger position of the gripper touches an object and applies sufficient force. To grasp objects with various profiles, the proposed granary soft gripper (GSG) is designed to be capable of enveloping, pinching and caging grasps. The gripper consists of a chamber palm, a palm cap and three fingers. First, the design of the gripper is analyzed. Then, after the theoretical model is constructed, finite element (FE) simulations are conducted to verify the built model. Finally, a series of grasping experiments is carried out to assess the snap-through behavior of the proposed gripper on grasping and sensing. The experimental results illustrate that the proposed gripper can manipulate a variety of soft and rigid objects and remain stable even though it undertakes external disturbances.

*Index Terms*—Soft gripper, Snap-through, compliant mechanism, Mechanical sensing, Robotic grasp.

## I. INTRODUCTION

GRASP is the most critical capability of robots, allowing robots to be deployed in practical scenarios[1-3]. The limitation of rigid grippers, owing to complex mechanical and control systems, has been overcome by soft grippers [4].
In particular, soft grippers offer better compliance and safe interactions among involved surroundings, which is especially capable of grasping soft objects[5].

In recent years, soft grippers have attracted considerable attention due to the advantages of the high compliance and robustness to variance in object geometry [6, 7]. Various emphasis on modeling and design, actuation technologies, materials, fabrication strategies, and so on[8]. Some soft gripper designs draw inspiration from the animals, such as human hands [8], elephant trunks [7], and starfish [9]. The soft grippers are constructed from soft functional materials, such as the dielectric elastomer [10] and the fluidic elastomer[11]. Direct 3D printing has offered a universal solution for fabricating soft grippers. Depending on the type of material used for the fabrication, different fabrication techniques, such as projection stereolithography[12], fused deposition modeling [13], are developed, allowing the gripper to be robust to environments.

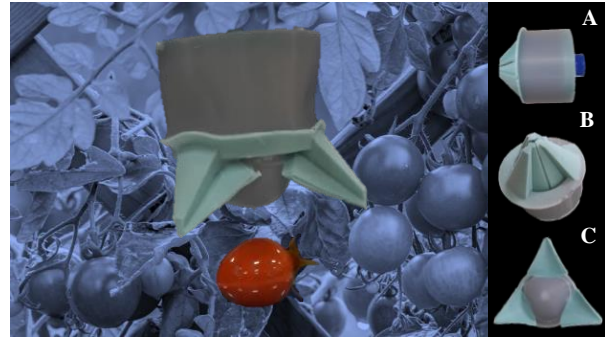

Figure. 1. The application scenario where the proposed GSG grasps small tomatoes. The three views of the gripper (A, B, C).

The most critical characteristic of compliant grippers is self-adaptive to the geometries of objects[14-16]. Compliant mechanisms, such as bistable buckled beams, are applied on soft robots. The two states of stable positions can be transformed under external disturbance, which is called a snap-through phenomenon. A sufficient force or bending moment allows the snap-through mechanism to be triggered[17]. The snap-through mechanism is commonly used for building micro-electro-mechanical systems (MEMS) [18, 19]. Snap-through characterizes a motion, which is applied in constructing mechanisms of climbing robots[20] and jumping robots[21], actuators and sensors[4, 22-25]. The heat, force, moment, electric field, magnetic field, and so on can be considered as the trigging sources of a snap-through mechanism[26]. Rothemund et al. [23] introduced a soft, elastomeric valve actuator based on the snap-through mechanism to control airflow, allowing for the rapid transition of different states[23]. The bistable snap-through mechanism was applied to constructing dielectric actuators since it can readily generate a large displacement [27]. A few bistable grippers based on the snap-through mechanism have also been proposed over the years[4, 28]. For example, an artful design of the gripper with parallelogram bars achieves a bistable mechanism based on the snap-through phenomenon[24]. Follador et al.[25] proposed a gripper consisting of two double-bistable structures based on dielectric elastomer actuators. Thuruthel et al.[4] proposes a soft robotic gripper combing a ring and a cross-shape structure, which enable the gripper to have the bistable characteristics for realize a grasp without electronic sensors.

However, the gripper presented in [24] is limited to grasping soft objects owing to the lack of sufficiently compliant capability. The robotic hand is constructed by individual fingers and thus, it is not available to realize the high coordination, which results in requiring a complex control strategy[25]. The fully soft bistable gripper [3] is capable of soft sensor-less sensing, the gripper cannot be opened automatically. Thus, to employ the gripper in practical applications, a human has to be involved in opening the fingers. In addition, without the corresponding actuation, the gripper cannot adjust the holding force.

In this work, to solve the above challenges, we present a granary-like soft gripper that can perform enveloping, pinching and caging grasps of soft, brittle, fragile, light, or very heavy objects, achieving 'sensing' and 'actuation' by a compliant bistable snap-through mechanism, which is applied to picking up fruit in agricultural scenarios (see Fig.1). The proposed granary soft gripper (GSG) is actuated by a pneumatic actuator and consists of a chamber palm, a palm cap and three fingers. The soft gripper is fabricated by the molding method of two types of materials. The proposed snap-through-based structure can store strain energy that can then be readily released upon generating a sufficient deformation. The default status of the GSG is closed. Thus, in terms of open-close transmission, it is no surprise that there is no power being consumed while the gripper is holding a light object. In addition, a relatively better sealing at the default mode allows the GSG to be potentially used in dirty environments. Such a design enables the gripper to generate grasp action when the palm mechanically touches an object with a certain pressure, while the sensing is completely passive. The joint structures between the palm and fingers in the proposed GSG are considered as the actuation mechanisms that have similar functions to four-bar linkages. There is the other sensing mode based on the proposed snap-through mechanism. In particular, before reaching the critical equilibrium point of the close state, we also can allow this sensing to be active by setting the air pressure threshold. When a pressure valve reads the threshold, inhaling from the chamber palm brings the gripper to close. Unlike the gripper [3], the proposed GSG can also increase holding force for achieving a stable grasp through changing the air pressure. Experimental results illustrate that the proposed GSG can realize sensor-less, passive, closed-loop grasps of a various range of objects with a maximum weight of around 200g and a maximum diameter of about 54mm (spherical objects).

We organize the rest of the paper as follows. The construction of the GSG is introduced in detail in Section II. A series of grasping experiments is conducted for evaluating the capabilities of the GSG in section III. We make a conclusion in

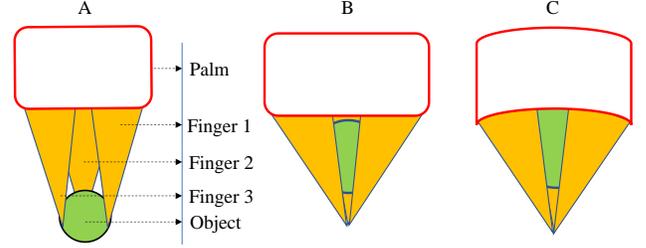

Figure. 2. The three grasp modes, such as the pinching grasp(A), enveloping grasp(B) as well as caging grasp(C).

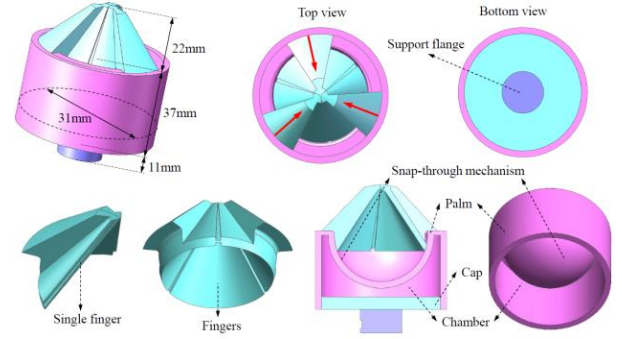

Figure. 3. The design of the proposed GSG.

Section IV.

## II. METHODOLOGY

In this section, we first study a generalized grasp mode of the proposed GSG and then construct the theoretical model based on the snap-through mechanism. Further, simulations are carried out for verifying the design. Finally, we describe the fabrication of the GSG.

### A. Design of the GSG

When attempting to grasp soft objects, we humans tend to use multiple fingers rather than just two fingers, which requires minimal effort for realizing a stable grasp, especially for grasping a spherical object. In terms of the detailed design, we need to consider a few factors, but the primary ones are (1) high payload capacity(0.2kg); (2) capability to stably grasp various objects (e.g., spherical, cubic, cylindrical, irregular geometrics, small, heavy objects). A gripper with three fingers is sufficient to achieve stable grasps for most objects to be grasped [29]. Thus, we design a soft gripper with three fingers so that a delicate object is less deformed compared with two-finger grasping. Moreover, it is easy for the three-finger gripper to achieve a caging grasp, which is a better form-closure grasp since it is not always possible to stably pick up some objects via enveloping and pinch grasps without a palm involved. The GSG has three grasping modes, such as pinching, enveloping as well as caging modes, as illustrated in Fig.2. The orange blocks represent soft fingers and the block formed by the red lines represent the palm. Here the proposed gripper with three fingers pushes an object lying on a platform by the palm against the object surface. A stable grasp is finally obtained as the palm and the fingers conduct covering action after the pressure force arrives at a threshold.

The detailed design of the GSG is shown in Fig.3. The GSG

is modeled as just one single component, consisting of a chamber palm, a palm cap and fingers in a symmetrical tri-fingered configuration to offer significant flexibility and versatility in tackling different profiles of objects. For such a design, three fingers with one single degree of freedom (DOF) are actuated simultaneously. The soft fingers can open and close with the palm chamber stretching, which is actuated by a pneumatic control system. The fingers are designed to be curved with a gradual thinner taper from the contact point and the finger ends for realizing a sealing state when the gripper is at the default mode. In addition, such a design for the fingers enables the palm touches an object ahead of fingers while the GSG keeps an opening state. The palm is also involved in increasing the contact areas of grasping an object, which improves grasp force and enables better contact for an object with an irregular shape to achieve high grasping stability.

The GSG's palm is designed as a mechanical sensor to perceive the touching between the gripper and a target. When the center of the palm touches an object, the mechanical deformation from the palm chamber allows the fingers to enclose the object based on the snap-through phenomenon. Further, to achieve a stable grasp, the positive pressure is brought to bear on the palm chamber for increasing the grasp/holding force. A thick joint is in the inner side of the palm and finger, which makes it inextensible yet flexible for bending along the backward and lateral directions. When the air pressure is applied to the palm chamber, a net bending motion is created to open the fingers due to the positive pressure and close the fingers owing to the negative pressure.

*B. Model Construction*

The snap-through phenomenon is illustrated in Fig.4. A system with two stable equilibrium states generates biostability that has two local minima of potential energy. Here we utilize the snap-through properties to construct a gripper with the capability of mechanical force sensing. Initially, a beam in a default configuration with the lowest potential energy (1st stable state) is compressed from an upward force. Subjected to constraints at the two ends, the beam is transited from the closing state to the opening state, with the upward force gradually increasing. During this period, the beam (what is the meaning of smart beam) is in an unstable state. At a certain time instant, the beam is suddenly snapped through to the second stable state with the other lowest potential energy (2nd stable state). Such a design based on the snap-through phenomenon brings three advantages. First, our manufacturing process leaves out the complexity of 3D printing, manually assembling and bonding, just employing the molding process owing to the simple structure. Second, the gripper is capable of mechanical force sensing. In particular, the palm of the gripper touches the object to cause the palm deformation due to the small strain energy, which results in the closing state through the snap-through mechanism. Third, the snap-through mechanism enables a gripper to reduce the complexities of the actuation mechanisms and control system.

The theoretical model of the deformation of snap-through behavior are investigated for estimating the dynamic effects on the response. In terms of the above design, the structure of the proposed gripper is considered as a snap-through structure of the thin-walled elastic spherical shell with bistable characteristics. To simplify the calculation, the snap-through structures in [4] and our work are considered as 2D arched beams. The work proposed in [4] makes an assumption that the two ends of a bistable beam with the snap-through phenomenon are fixed without rotations; however, the rotations at the ends arise while the gripper grasps an object. Without losing the generality, we construct a general theoretical model of the snap-through mechanism with rotations at the ends based on the 2D structural system using the co-rotational approach[30]. The core idea of co-rotational modeling is to separate the rigid-body motions of beam elements in the global coordinates from their deformations in the local coordinates, thus making it able to model the arbitrarily large deflection of an object undertaking significant external loads or displacements, which is suitable for modeling the snap-through phenomenon.

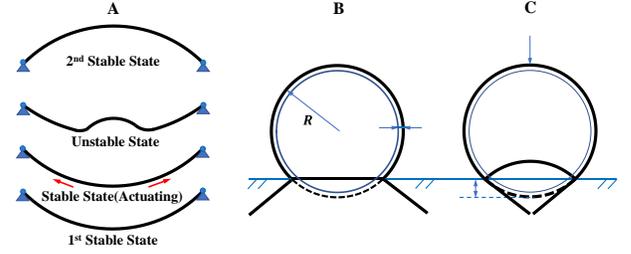

Figure. 4. The bistable snap-through phenomenon(A). The deformed configurations of the thin-walled spherical shells for the initial status(B) and the snap-through behavior(C).

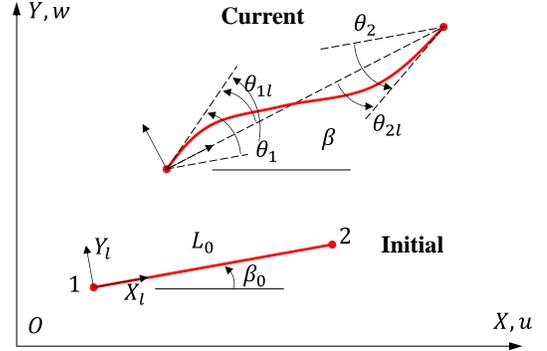

Figure 5. Relationship between global coordinates and local coordinates of the nodes of each beam element.

In terms of each beam element, it includes two ends such as node 1 and node 2, as shown in Fig.5. In the reference frame of the global coordinate $\{O, X, Y\}$, the coordinates of node 1 and node 2 are $(X_1, Y_1)$ and $(X_2, Y_2)$ at the initial configuration, respectively. Then, the initial angle and length of the beam are denoted by $\beta_0$ and $L_0$, correspondingly. When the beam undertakes an external load, it will be changed to the current configuration where two nodes have displacements $(u_1, w_1)$ and $(u_2, w_2)$ for arriving at the current positions with the current incline angle $\beta$ and length $L$ that can be derived from nodes' coordinates. As for the beam, its axial deformation $u_l$ can be achieved from the length change from $L_0$ to $L$. Thus, we can derive the axial force $F_N$ along the beam as follows

$$dX = (X_2 + u_2) - (X_1 + u_1)$$
$$dY = (Y_2 + w_2) - (Y_1 + w_1) \quad (1)$$
$$L_0 = \sqrt{(X_2 - X_1)^2 + (Y_2 - Y_1)^2}$$

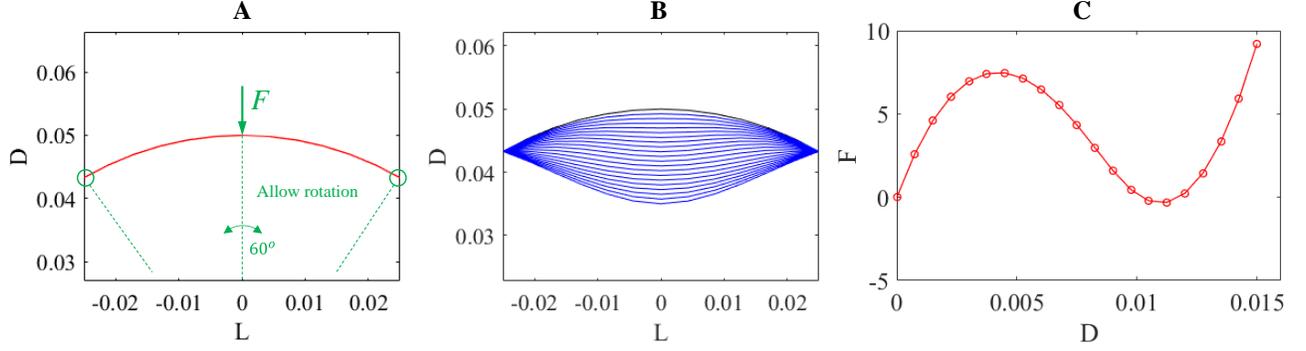

Figure. 6. The Force applied to the beam with the rotations at the two ends(A); the deformations of the snap-through mechanism with the time varying(B); The forces and deformations for the snap-through behavior(C). $D(m), L(m)$ and $F(N)$ represent the deformation, length and applied force, respectively.

$$L = \sqrt{(dX)^2 + (dY)^2} \quad (2)$$

$$u_l = \frac{L^2 - L_0^2}{L + L_0} \quad (3)$$

$$\cos\beta = \frac{dX}{L}, \sin\beta = \frac{dY}{L} \quad (4)$$

$$F_N = \frac{EAu_l}{L_0} \quad (5)$$

where $E$ and $A$ represent Young's modulus and cross-sectional area, respectively. Moreover, the beam element generates rotational motions at two nodes. The rotation angles $\theta_1$ and $\theta_2$ of node 1 and 2 can be measured according to the initial incline axis of the beam element (see Fig.5). Combining a node's linear displacement $(u, w)$, the global displacement of each node can be represented by the vector $(u, w, \theta)$. Thus, the local rotations can be provided as

$$\theta_{1l} = \theta_1 + \beta_0 - \beta$$
$$\theta_{2l} = \theta_2 + \beta_0 - \beta \quad (6)$$

Employing the standard structural system, the moments of the beam's nodes can be derived via

$$\begin{Bmatrix} M_1 \\ M_2 \end{Bmatrix} = \frac{2EI}{L_0} \begin{bmatrix} 2 & 1 \\ 1 & 2 \end{bmatrix} \begin{Bmatrix} \theta_{1l} \\ \theta_{2l} \end{Bmatrix} \quad (7)$$

in which $I$ is the moment of inertia. As a result, the global displacement $(u, w, \theta)$ of the beam can be applied to obtaining the local displacement $(\theta_{1l}, \theta_{2l}, u_l)$, and the latter can be further used for calculating the applied load $(M_1, M_2, F_N)$ through Eqs. (1-7). The load-displacements are illustrated in Fig.6 where two quasi-static states can be observed.

Figure 7 illustrates the finite element (FE) simulations of the constructed GSG regarding the deformations and forces to verify the proposed theoretical model. The beam is configured along the vertical direction and the applied force is executed in two steps. First, only its gravity is considered in the finite element model. The applied force is imposed on the beam(palm) to bring fingers to the open state during the second stage and the corresponding simulations on the deformations and equivalent stresses inside of the beam are illustrated in Fig.7. From the results, we found that the deformations of the beam during the period of arriving at the next stable state have almost similar changing trends with results provided in Fig.6, which indicates the constructed model is in line with forecast.

*C. Fabrication of the GSG*

We fabricate this gripper that can realize snap-through behaviors based on the proposed modularized design that significantly improves the robustness of the GSG. The gripper is fabricated with Smooth Sil 960 (Smooth-On, USA, Shore 60A) for the fingers and cap, and Dragon Skin 30 (Smooth-On, USA, Shore A hardness of 30) for the palm since the fingers need to have a good performance at preventing objects from slipping.

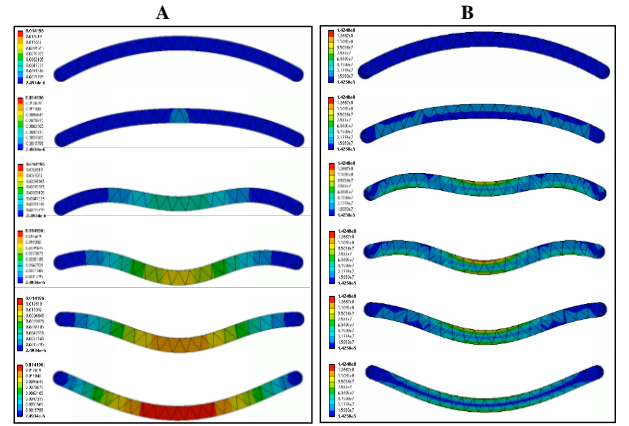

Figure.7. The deformations(A) and the equivalent stresses(B) of snap-through phenomenon based on the finite element analysis.

It is in general difficult and expensive for traditional 3D-printing, assembling and bonding approaches to fabricate soft robots. In this work, the molding method is applied to making the proposed GS gripper. The major fabrication procedure is illustrated in detail in Fig.8, with relevant steps labeled. The GSG fabrication begins with designing the molds that are applied to building the proposed GS gripper. There are three main stages for this fabrication. In terms of the first stage, these molds are 3D printed from a silicone elastomer by a MakerBot Replicator 2 (MakerBot Industries, LLC). After assembling these molds, we pour the mixed liquid silicone rubber (Smooth Sil 960, Smooth-On, USA) into the assembled mold that consists of three molds from a port. It is noted that the operations require to be careful and slow for avoiding air bubbles when the mixed liquid silicone rubber is poured into the 3D-printed molds. After putting them into an oven at 35°C for 24 hours to solidify the soft material, we open the assembled mold to obtain the palm. As for the second stage, the palm is also regarded as a mold for fabricating fingers (see Fig.8). The steps are similar to the above stage. The poured liquid silicone rubber is changed to Dragon Skin 30. It is noted that the three fingers are made as a whole part with three thin walls that are easy to be separated. In the third stage, the blue fiber-reinforced structure with six holes, as shown in Fig.8, can enhance the stiffness of the cap by being installed into the cap in case that the stretching of the palm causes large deformation of the cap as a contacting flange. After pouring, we considered the palm chamber as a mold and cover it on the bottom mold tightly. Further, these assembled molds are placed again into an oven with the same time and temperature mentioned above. Finally, we take the soft gripper out of the molds after the cured skeletons are removed from the molds. All the molds and components of the proposed GS gripper are assembled and disassembled, as shown in Fig.8.

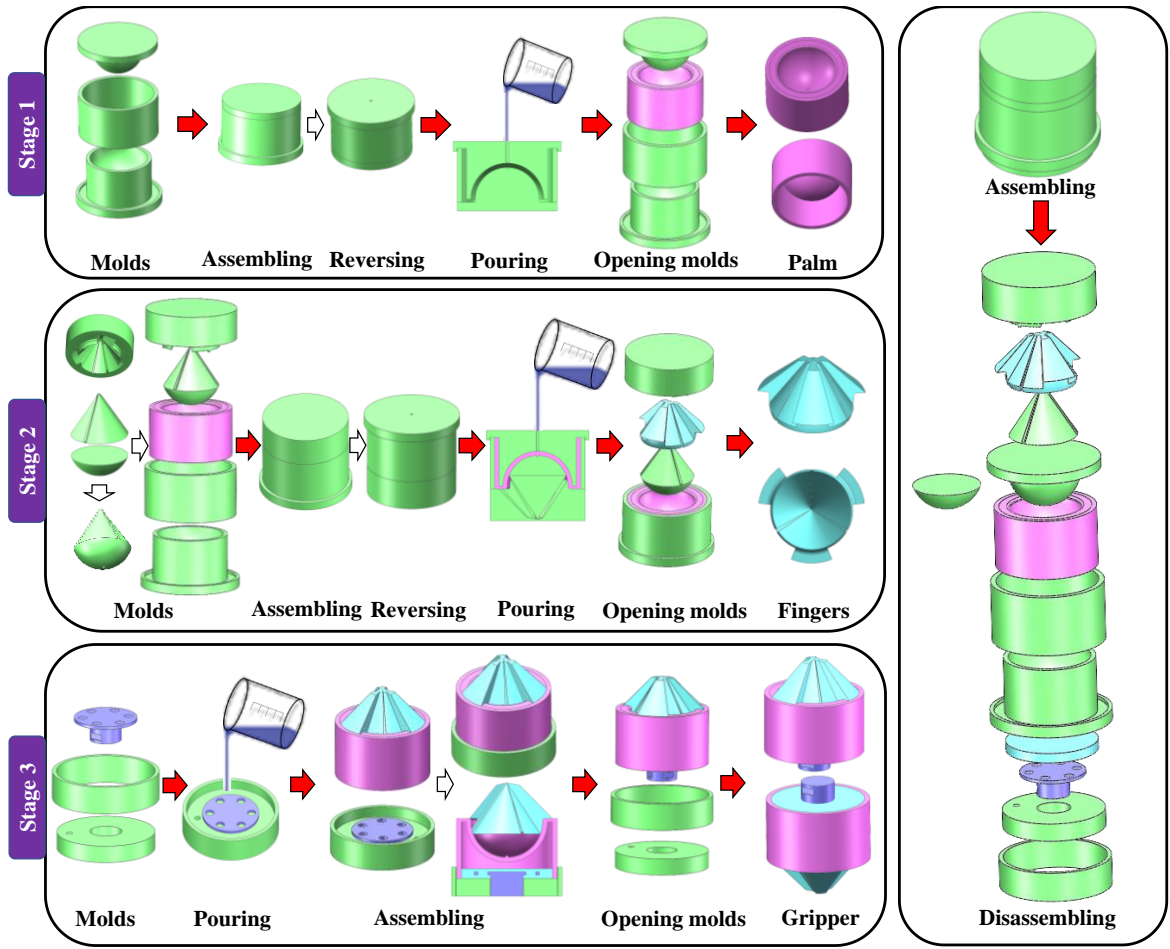

Figure.8. The fabrication stages of the proposed GS gripper. The first stage introduces how to mold the gripper's palm; the second one describes the fabrication process of the fingers; the third stage illustrates to obtain the gripper by opening the molds.

## III. Experiments

In this section, we first demonstrate the grasp behaviors of the proposed GS gripper based on the snap-through mechanism via a detailed analysis for the gripper grasping an object. Then, to evaluate the grasping performance of the gripper, we implement a series of grasping tasks that show the full functions and potential agricultural applications of gathering fruits.

### A. Gripper Behaviors

The holding force of the proposed GS gripper is investigated. It is well-known that two factors, including target geometry and palm actuation, contribute to characterizing gripper force. We printed two objects, including a sphere with a 30mm diameter and a cube with a 30mm size, for the experiments. The object was firmly attached to the load cell so that an external force could be applied to pulling the gripper until the grasped object was released. The readings of the pulling force were recorded during the pulling period. The peak holding force was again achieved while the gripper held the sphere, reaching approximately 10N. For the holding force concerning the cube, it is significantly smaller than that of the sphere since three fingers and a soft palm forms a cage to prevent the sphere from sliding out, with the same vacuum pressure applying.

The open-close action source of the proposed GSG is from the snap-through mechanism of the palm. We used a valve vacuum sensor to approximately record the pressure of the palm chamber, as shown in Fig.9. In terms of the closing process (see Fig.9-A), at the initial phase, the pressure magnitude smoothly increased with the fingers closing. At a moment, the fingers reached the stable status so that the required vacuum pressure magnitude became small. The snap-through characteristics resulted in the pressure saltation. To close fingers tightly, the big pressure magnitude was applied to the palm. The default state of the proposed GS gripper is closed. For the opening process, the pressure frequently changed until the two fingers opened to arrive at a stable status. Figure 9(B) illustrates that there are two pressure peaks, such as 13 and 11, of the palm chamber before fully opening since three fingers reached stable status at different time samples. The manually molding method results in the difference between the two jointing structures (between the fingers and palm) and the third one. This also indirectly verifies that the experimental results have a good agreement with the previous theoretical model.

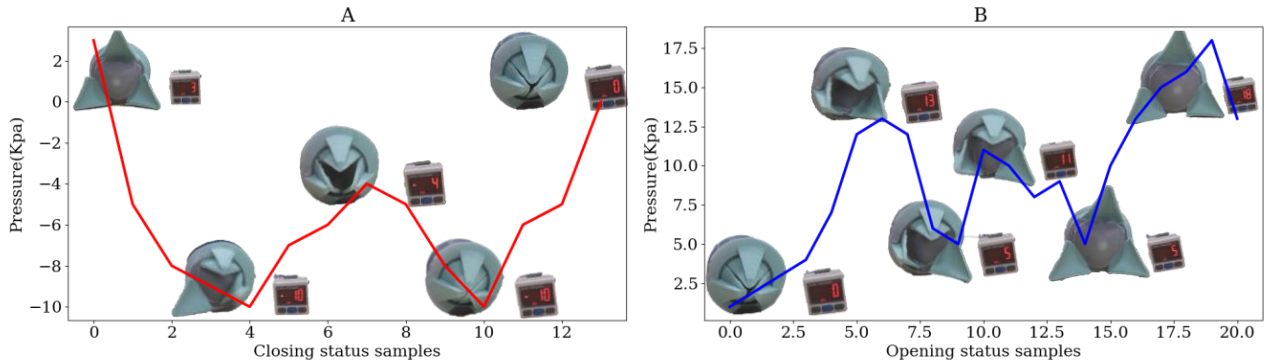

Figure.9. The pressures of the closing status(A) and the opening status(B) with the corresponding samples.

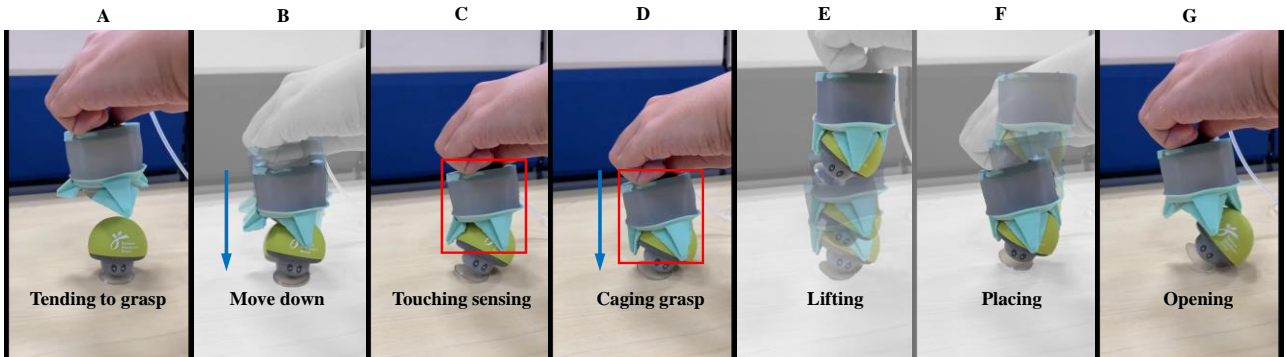

Figure.10. The detail process of the proposed GS gripper grasping an object with a relatively big size. The blue arrow represents the moving direction and the red frames denote the current positions of the proposed GSG, respectively.

The experimental grasping system includes the proposed gripper, the vacuum actuation system and the vacuum pressure sensor. Thus, the GSG could conduct two grasping modes based on the proposed force-sensing method, such as the passive and active modes. In terms of the passive mode, the palm touches the target to continue moving down; when the deformation of the palm reaches a threshold status that causes the snap-through phenomenon, the fingers passively close to grasp the target object. In addition, we set a vacuum threshold in the control system to enable the grippers to actively close. Specifically, the palm has contact with the target object. If the contact force is larger than the set threshold, the fingers will close to grip the object after a valve pressure sensor detects the palm pressure. Figure 10 illustrates the detailed process of the GSG grasping an object from the gripper approaching status to the successful lifting status.

### B. Grasp Evaluation

The GSG was applied to picking up a wide variety of objects spanning different geometries, and surface finishes and payloads. The GSG was able to grasp delicate sphere-like items such as small tomatoes, grapes, and cherries without breaking or damaging them. In addition, the GSG also could grasp irregular-shaped objects such as a maize cob, medicine bottle and a box. The details of all the objects with different weights, geometries and surface materials are presented in Table I. We conducted the grasp experiments to evaluate the GSG's repeatability in grasping tasks, as illustrated in Fig.11. The proposed GS gripper attempted to grasp the object along the direction that is perpendicular to the table. To illustrate the performance of stable grasps, the proposed gripper swayed each object in three times without dropping it, which is considered a successful grasp. The inflating or releasing of the GSG was manually controlled.

Table I. The dimensions and weights of objects. D(mm) and W(g) represent the dimension and weight, respectively.

| Items | D | W | Items | D | W |
|---|---|---|---|---|---|
| Motor box | 2 | 17.6 | Grape | 27 | 2.3 |
| Cartridge | 5 | 2 | Weight | 28 | 200 |
| Jujube | 12 | 1.5 | Small tomato | 29 | 7 |
| Glue | 17 | 15.9 | Cherry | 31 | 1.3 |
| USB driver | 18 | 9.2 | Strawberry | 37 | 3.5 |
| Medicine bottle | 18 | 12.2 | Maize cob | 45 | 57 |
| Plum | 24 | 1.6 | Speaker | 54 | 69.2 |

To obtain the maximum weight and size of the object to be successfully grasped, we perform two groups of grasping experiments. As for the first group, the weight of the object with the initial value 100g was gradually increased by 20g every time after the gripper accomplishes a few pick-and-place actions for each object. Through repeating many grasp trails, we conclude that the proposed gripper could reliably grasp a range of targets with weights reaching nearly 200g (the weight information in deep gray color mentioned in Table I), as shown in Fig.11. Similarly, we conduct the second experimental group to achieve a range of objects' sizes. The maximum diameter of the object was 54mm that is approximately 174% of the GSG's diameter (31mm) (see the speaker information in black color at

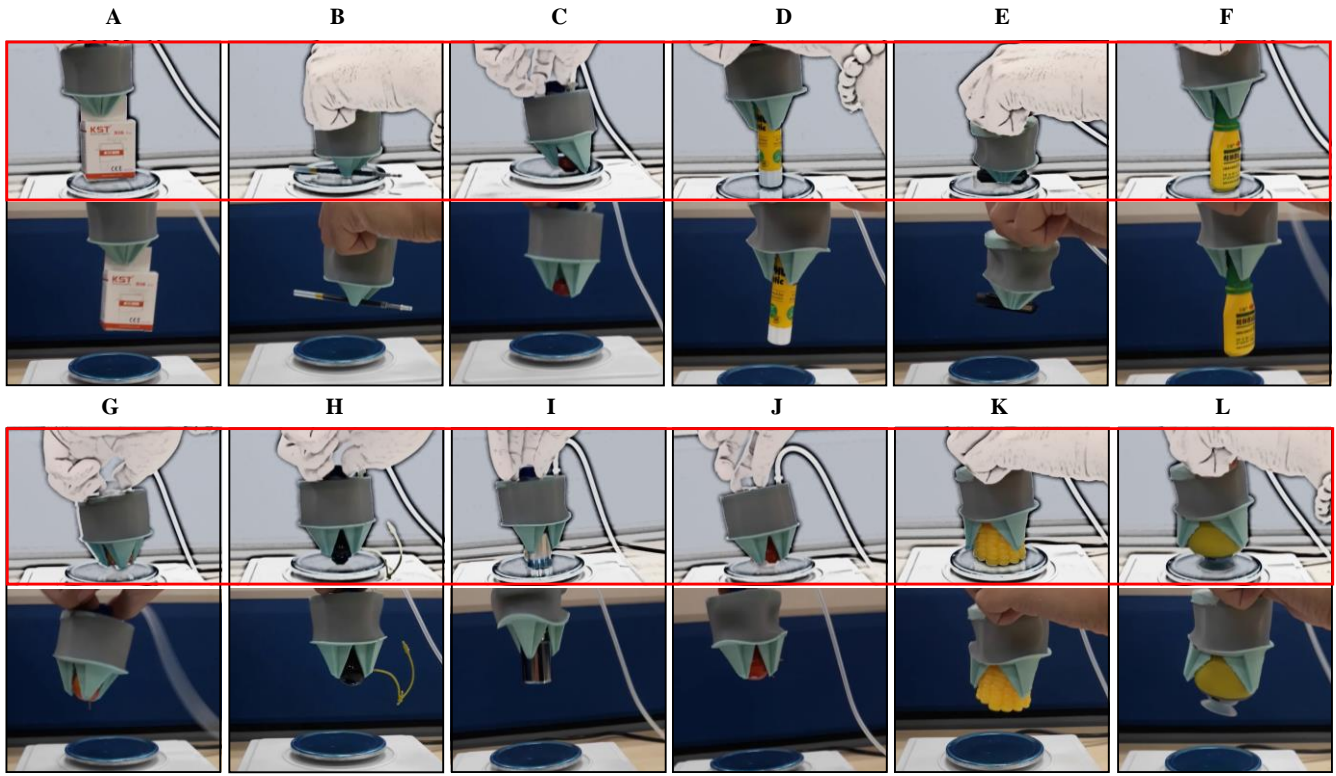

Figure. 11. The proposed GS gripper grasps the objects including Motor box(A), Cartridge(B), Jujube(C), Glue(D), USB driver(E), Medicine bottle(F), Plum(G), Grape(H), Weight(I), Small tomato(J), Maize cob(K) and Speaker(L). For each item, the top and bottom pictures show the gripper moves towards to grasp and lifts up the object steadily, respectively.

Table I). The proposed GSG could perform pinching grasps for small objects, enveloping grasps for objects with big sizes and caging grasps for small sphere-like objects. For instance, the pinching grasp tests were carried out for demonstrating how the GSG could manipulate objects with small profiles, which is not specially designed for sphere-like objects, as shown in Fig.11(A, B). In terms of sphere-like objects (Jujube, Plum, Grape, Small tomato, Cherry), the gripper performed enveloping and caging grasps. Here we define a caging grasp that the palm and all the fingers are involved in covering the object. Although some objects' contours fit poorly inside the gripper's grasping diameter, the proposed GSG can still be conformed to irregular shapes, allowing to grasp them such as the maize cob and speaker(see Fig.11). In addition, when performing a caging grasp, the gripper tended to move down for allowing the object to enter the space of the palm to achieve a stable grasp.

The proposed GSG is particularly well-suited for grasping sphere-like objects, soft and delicate objects due to their compliant skeleton and caging shape factor; however, it is in general not available for a gripper with two fingers to manipulate such spherical objects since two fingers of the gripper readily squeeze out a spherical object. It is found that the gripper tended to perform caging gasps for heavy objects. The caging grasp offers a more stable grasp than enveloping, pinching grasps. In this case, the contact surface is adequately big among the object, fingers and palm, which allows the grasping force to be distributed based on the design of the GSG; otherwise, the object would be easily damaged or be dropped.

## IV. CONCLUSION

In this work, we constructed a novel granary-like soft gripper based on the proposed snap-through structure that provides the actuation mechanism and sensor-less perception. Various experiments were carried out to verify that the proposed GSG is capable of grasping a variety of rigid, soft, delicate objects with different profiles while being gentle enough to avoid causing damage. In terms of future work, we will focus on how to optimize the soft gripper with the snap-through properties.

## ACKNOWLEDGMENT

This work was supported by Agency for Science, Technology and Research, Singapore, under the National Robotics Program, with A*star SERC Grant No.: 192-25-00054 and Robotics Enabling Capabilities and Technologies with Grant No.: R-397-000-381-305.

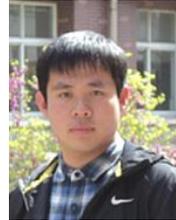

**Huixu Dong** (S'17-M'18) received the B.Sc degree in mechatronics engineering from Harbin Institute of Technology in China, in 2013 and obtained Ph.D. degree at Robotics Research Centre of Nanyang Technological University, Singapore 2018. He was a post-doctoral fellow at Robotics Institute of Carnegie Mellon University and currently, is a research fellow at National University of Singapore. His current research interests include robotic perception and grasp in unstructured environments, mechanism design with new concepts.

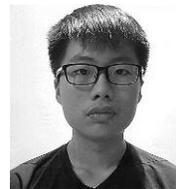

Chao-Yu Chen received his Bachelor in Mechanical Engineering from National Taiwan University in 2018. He is currently a master candidate in Biomedical Engineering from National University of Singapore and a research engineer with the Evolution Innovation lab at Advanced Robotics Centre. Chao-Yu is also a co-found of a soft robotics startup founded in 2021.

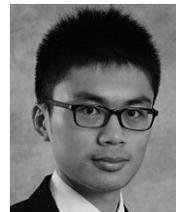

Chen Qiu received the B.S. degree from Beihang University, China, in 2011, and then the Ph.D. degree at the Centre for Robotics Research, King's College London, U.K in 2016. He was a research fellow of Robotics Research Center of Nanyang Technological University, Singapore. Currently, he is the director of the research department Maider Medical Industry Equipment Co., ltd. His research interests include soft robotics and medical robots.

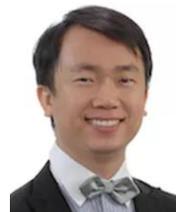

Raye CH Yeow received his PhD in Bioengineering from the National University of Singapore in 2010, and his post-doctoral training with the BioRobotics Laboratory, Harvard University from 2010-2012. He is an Associate Professor and a Principal Investigator with the Advanced Robotics Centre. Dr. Raye received the MIT Technology Review Innovators Under 35 Asia in 2016, and the Applied Research Award, Young Creator Award, and Technology Innovation Award from the IES Prestigious Engineering Awards 2017-2018. He is also a co-founder of two soft robotics startups.

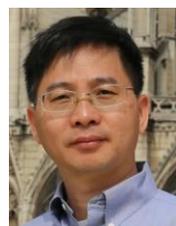

Haoyong Yu received the B.S. and M.S. degrees in Mechanical Engineering from Shanghai Jiao Tong University, Shanghai, China, in 1988 and 1991 respectively. He received the Ph.D. degree in Mechanical Engineering from Massachusetts Institute of Technology, Massachusetts, USA, in 2002. He was a Principal Member of Technical State at DSO National Laboratories, Singapore, until 2010. Currently, he is an associate professor of Advanced Robotic Centre and the department of biomedical engineering at the National University of Singapore. His current research interests include biomedical robotics and devices, rehabilitation engineering and assistive technology, biologically inspired robotics, intelligent control and machine learning.